\pdfoutput=1
\documentclass[runningheads]{llncs}
\usepackage{graphicx}
\usepackage{caption} 
\usepackage{cite}
\usepackage{latexsym}

\usepackage{url}
\usepackage{hyperref}

\usepackage{amsmath}
\usepackage{amssymb}
\usepackage[linesnumbered,ruled]{algorithm2e}
\usepackage{booktabs} 
\usepackage{multirow}
\usepackage{caption}
\usepackage{graphicx}

\usepackage{numprint}
\usepackage[british]{babel}

\usepackage{array}
\newcolumntype{L}[1]{>{\raggedright\let\newline\\\arraybackslash\hspace{0pt}}m{#1}}
\newcolumntype{C}[1]{>{\centering\let\newline\\\arraybackslash\hspace{0pt}}m{#1}}
\newcolumntype{R}[1]{>{\raggedleft\let\newline\\\arraybackslash\hspace{0pt}}m{#1}}

\begin{document}
%
\title{Cross-Lingual Transfer for Distantly Supervised and Low-resources Indonesian NER}
%
%
\author{
    Fariz Ikhwantri\inst{1}
}
    %
    %
\institute{
    Kata Research Team, Kata.ai \\
    \email{fariz@kata.ai}\\
}
\maketitle              
\begin{abstract}

Manually annotated corpora for low-resource languages are usually small in quantity (gold), or large but distantly supervised (silver). Inspired by recent progress of injecting pre-trained language model (LM) on many Natural Language Processing (NLP) task, we proposed to fine-tune pre-trained language model from high-resources languages to low-resources languages to improve the performance of both scenarios. Our empirical experiment demonstrates significant improvement when fine-tuning pre-trained language model in cross-lingual transfer scenarios for small gold corpus and competitive results in large silver compare to supervised cross-lingual transfer, which will be useful when there is no parallel annotation in the same task to begin. We compare our proposed method of cross-lingual transfer using pre-trained LM to different sources of transfer such as mono-lingual LM and Part-of-Speech tagging (POS) in the downstream task of both large silver and small gold NER dataset by exploiting character-level input of bi-directional language model task.

\end{abstract}
\section{Introduction}
Building large named entity gold corpus for low-resource languages is challenging because time consuming, limited availability of technical and local expertise. Thus, manually annotated corpora for low-resource languages are usually small, or large but automatically annotated. In most cases, the former are used as a test set to evaluate models trained on the latter one.

To reduce the annotation efforts, previous works \cite{P17-1135} utilized parallel corpus to project annotation from  high-resource languages to low-resources languages using word-alignment. Another promising approach is to use knowledge base e.g DBPedia \cite{Alfina2017ModifiedDE, Alfina2016DBpediaEE} or semi-structured on multi-lingual documents e.g Wikipedia \cite{U08-1016} to generate named entity seed. 

Previous works on multi-lingual Wikipedia with motivation to acquire general corpus \cite{U08-1016} and knowledge alignment between high--resource and low--resource languages encounter low recall problem because of incomplete and inconsistent alignments \cite{P17-1178}. Some work on monolingual data with intensive rule labelling \cite{Alfina2016DBpediaEE} and label validation \cite{Alfina2017ModifiedDE} to create automatic annotation also face the same problem.

Our contribution in this paper consists of two parts. First, we propose to improve NER performance of a low-resource language, namely Indonesian, trained on noisily annotated Wikipedia data by (1) fine-tuning English NER model, and (2) using contextual word representations derived from either English (EN), Indonesian (ID), or Cross-lingual (EN to ID) fine-tuning of pre-trained language models which exploit character-level input. Second, we analyze why using pre-trained English language model from \cite{Peters:2018} yields improvement compare to monolingual Indonesian language model by looking at the dataset size, shared characteristic such as orthography, and its different like grammatical and morphological different to source language (English). We show that fine-tuning ELMo in unsupervised cross-lingual transfer can improve the performance significantly from baseline Stanford-NER \cite{P05-1045}, CNN-LSTM-CRF \cite{P16-1101} and previous works using state-of-the-art multi-task NER with language modeling as an auxiliary task \cite{P17-1194, W18-6112} trained on conversational texts, and its monolingual counterpart that is trained on different dataset size in the target language, which in our case is Indonesian unlabeled corpora retrieved from Wikipedia and news dataset \cite{tala2003study}.

\section{Related Works}

Recently, Peters et al, \cite{Peters:2018} proposed to use pre-trained embedding from language model (ELMo) of large corpora for many NLP tasks such as NER \cite{TjongKimSang}, semantic role labeling \cite{Palmer2005ThePB}, textual entailment \cite{Bowman2015ALA}, question answering \cite{D16-1264} and sentiment analysis \cite{D13-1170}. Motivated by deep character embedding for word representation that is useful in many linguistic probing and downstream tasks \cite{Perone2018EvaluationOS} and trained on large corpora using language model objective, we chose to investigate ELMo embedding as weight-initialization for NER task in a low-resource languages. 

\subsection{Deep Character Embedding}


Character embedding is important to handle out-of-vocabulary problem such as in out-of-domain data \cite{W18-6112} or another language with shared orthography \cite{I17-2016}. The input words to Bidirectional LM, are computed by using concatenation of multiple convolution filters over sum of characters sequences of length \cite{Kim:2016:CNL:3016100.3016285, 45446}, 2 depth highway layers \cite{srivastava2015highway} and a linear projection.

The input to highway layers $y_{k}$ is the concatenation of $y_{k,1},...,y_{k,h}$ from $H_{1},...,H_{h}$ as $y_{k}=[y_{k,1},...,y_{k,h}]$.
The output $x_{h}$ of highway layers of depth $h$ are computed as in Equation \eqref{eq:highway}, where $T = \sigma (W_{\textit{T}}x_{\textit{h}-1}+b_{\textit{T}})$ and, $x_{0} = y_{k}$ as an input to the first highway layer.

\begin{equation}\label{eq:highway}
        x_{\textit{h}} = T \odot ( W_{\textit{H}} x_{\textit{h}-1}+b_{\textit{H}}) + (1-T) \odot x_{\textit{h}-1} 
\end{equation}

\subsection{Bidirectional Language Models (BiLM)}\label{subsec:lm-objective}

Language modeling (LM) computes the probability of token $t_k$ in sequence of tokens length N given the preceding tokens $(t_1,t_2,...,t_{k-1})$ as $\log p(t_1,t_2,...,t_N)=\sum_{k=1}^{N} \log p(t_{k}|t_1,t_2,...,t_{k-1})$. Reversed order LM, computes the probability of token $t_k$ in a sequence of tokens of length $N$ given the succeeding tokens in $\log p(t_{k+1},t_{k+2},...,t_{N})$ as  $p(t_1,t_2,...,t_N)=\sum_{k=1}^{N} \log p(t_{k}|t_{k+1},t_{k+2},...,t_{N})$.

\begin{equation}\label{eq:bilm}
\begin{split}
    \sum_{k=1}^{N} ( \log p(t_{k}|t_1,t_2,...,t_{k-1}| \theta_{x}, \overrightarrow{\theta}_{LSTM}, \theta_{s}) \; + \\ \log p(t_{k}|t_{k+1},t_{k+2},...,t_{N}| \theta_{x}, \overleftarrow{\theta}_{LSTM},\theta_{s}))
\end{split}
\end{equation}

In downstream task such as NER sequence labeling, the output of ELMo \cite{Peters:2018} used for contextual word representation is the concatenation of projected highway layer \cite{srivastava2015highway} of Deep Character Embedding output \cite{Kim:2016:CNL:3016100.3016285, 45446}, forward and backward output of LM-LSTM output of hidden layer. There are several ways to use ELMo layer for sequence labeling task, one of them is to use only last layers output of BiLM-LSTM. In this research, we only explore using last hidden layer of BiLM-LSTM \cite{P17-1161}. 

\subsection{Cross-lingual Transfer via Multi-Task Learning}

Cross-lingual transfer learning aims to leverage high--resources languages for low-resource languages. Yang et al., (2016) \cite{Yang2016TransferLF} proposed to transfer character embedding from English to Spanish because they shared same alphabet, while Cotterell et al., (2017) \cite{I17-2016} study several languages transfer within the same family and orthographic representation using character embedding as shared input representation. In their proposed model, they shared character convolutions for composing words but not the LSTM layer. In the previous works above, the training process minimizes the joint loss of low-resource and high-resource languages as supervised multi-task learning (MTL) objective. However we found that due to grammatical and morphological different, it is more significant to do pre-training scenario (INIT) instead of joint-training objective.

\section{Proposed Method}

\begin{figure*}[htbp]
  \centering
    \includegraphics[width=1.0\textwidth]{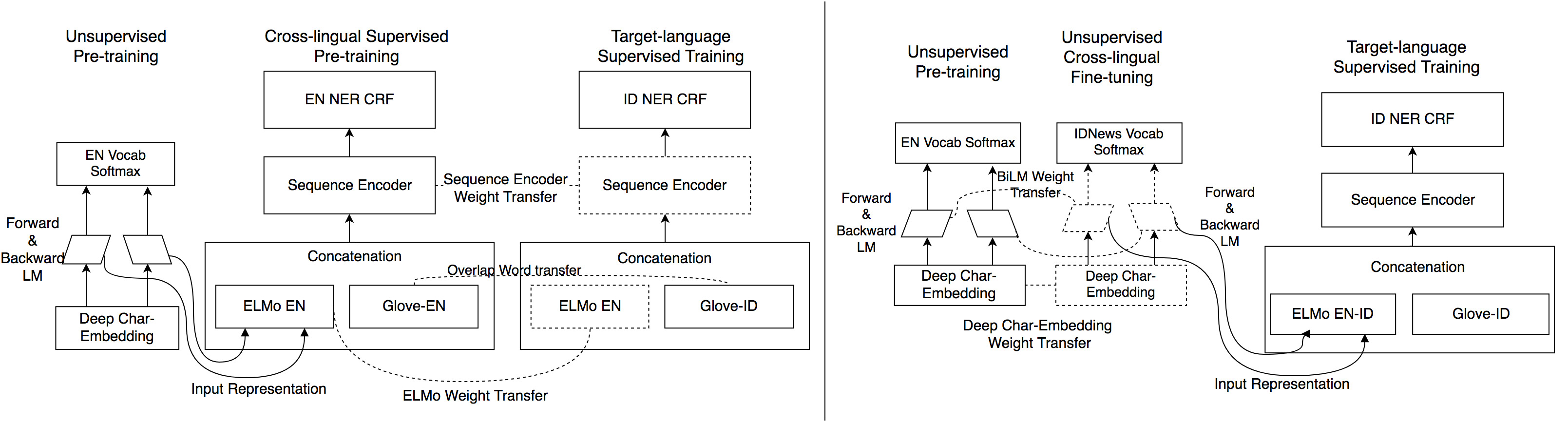}
    \caption{Cross-lingual Transfer Learning by using Character-level pre-training. Left image, our proposed Unsupervised-Supervised Cross-lingual Transfer where we fine-tune ELMo on target task NER but on source language. Right image, our proposed Cross-lingual Language Model fine-tuning where we fine-tune ELMo on target language Indonesian}
    \label{fig:uss-uus}
\end{figure*}

In this section we explain briefly our two proposed method. Our first proposed method extend supervised cross-lingual transfer using ELMo (Figure \ref{fig:uss-uus}, left image). Our second proposed method fine-tune ELMo from English to Indonesian News dataset to use on distantly supervised and small gold Indonesian NER dataset.

\subsection{Supervised Cross-lingual Transfer with ELMo}\label{subsec:USS}

Alfina et al \cite{Alfina2017ModifiedDE} observed that automatically annotated corpora fail to tag many orthographically similar entity of "America"  to "Amerika" in Indonesian. We also confirmed that, there are many cases of false negative in orthographically similar \texttt{LOCATION}  alias such as "Pacific" to "Pasifik" in Indonesian Wikipedia. Intuitively, we proposed to increase the recall performance due to many false-negative error by supervised cross-lingual transfer \cite{Yang2016TransferLF} using pre-trained weights from state-of-the-arts NER model that uses Bidirectional language model. In the experiment result Table \ref{tab:exp-result}, the model corresponds to [English NER Sources] \texttt{ELMo EN-1B Tokens} from "Supervised CL Transfer with ELMo" scenario.

\subsection{Unsupervised Cross-lingual Transfer via ELMo fine-tuning}\label{subsec:UUS}

We proposed to use a pre-trained language model of high-resource languages such as English in order to initialize better weights for low-resource languages. The cross-lingual transfer in our research is simple and almost the same as \cite{P18-1031} with language modeling objectives but we replace English target vocab with Indonesian by random initialization (figure. \ref{fig:uss-uus}, right image). 

Our motivation to propose this method is because we observed that there are only marginal improvement using monolingual Indonesia LM of 82M Tokens from Wikipedia compared to using English LM trained on 1B Tokens on applying ELMo to Distantly Supervised NER dataset. This might be attributed due to large difference of publicly available unlabeled corpus size, such as 82M in Indonesia Wikipedia\footnote{as of 20-08-2018 \href{https://dumps.wikimedia.org/idwiki/20180820/}{Wikipedia Database dump}} vs 1B Tokens of language model benchmark or 2.9B English Wikipedia available to train. In the experiment result Table \ref{tab:exp-result}, the model corresponds to \texttt{ELMo EN-ID Transfer} from one of the "\texttt{CL via ELMo EN}" group scenario.

\section{Dataset}

In this research, we used gold and silver annotation named entity corpus in English as sources in transfer learning. For target language, we used large silver annotation Indonesian as training dataset. We use two set of small clean $< 40$k tokens and $\leq 1.2$k sentences as testing data in model comparison scenarios and another one as training data in ablation scenario for analysis, in addition of unlabeled data from Wikipedia and newswire.

\subsection{Gold named entity corpus}

\subsubsection{CoNLL 2003}
Dataset is well known shared task benchmark dataset in many NLP experiment. We follow the standard training, validation (testa), and test (testb) split scenario. The label consist of \texttt{PERSON}, \texttt{LOCATION}, \texttt{ORG}, and \texttt{MISC}. We experiment additional scenarios for cross-lingual transfer which ignore \texttt{MISC} labels.

\subsubsection{Clean 1.2K DBPedia}
Human annotations for a subset of the silver annotation corpus are important to measure the quality of that automatic annotation. Thus, we asked an Indonesian linguist to re-label the subset of data and compute the metrics for DEE, MDEE and +Gazz silver annotation dataset. The precision, recall and F1 score of the subset w.r.t our clean annotation can be found in Table \ref{tab:silver-data}. The clean annotation can be found at \href{????}{data supplementary material}. We used this in-house annotation to do ablation analysis after training distantly supervised NER. We will made this subset of cleaned DBPedia Entity from noisy annotation publicly available in order to allow others to replicate our results in low-resources (gold) scenario.

\subsection{Noisy named entity corpus}

\subsubsection{Wikipedia Named Entity}

WP2 and WP3 are two version of dataset \cite{U08-1016}. The corpus obtained from this github repository\footnote{https://github.com/dice-group/FOX/tree/master/input/Wikiner}, because the initial link mentioned in the \cite{U08-1016} is down.  In this research we use these 2 version that corresponding to WP2 and WP3 of this silver standard named entity recognition dataset. We evaluate this dataset on CoNLL test \cite{TjongKimSang} and WikiGold \cite{Balasuriya:2009:NER:1699765.1699767}. 

\begin{table}
\parbox{.5\linewidth}{
    \centering
    \footnotesize
    \caption{Dataset statistics used in our experiments. \#Tok: numbers of tokens. \#Sent: numbers of sentences. Alfina et. al. \cite{Alfina2016DBpediaEE, Alfina2017ModifiedDE} use \textbf{Gold as their test set}. Clean 1.2K are used to measure noisy percentage of DEE, MDEE, and +Gazz and low-resources scenario}
    \begin{tabular}{l @{}|c|c|c|c|r@{}}
        \toprule
        Dataset & PER & LOC & ORG & \#Tok & \#Sent \\
        \midrule
        DEE & 13641 & 16014 & 2117 & 599600 & 20240 \\
        MDEE& 13336 & 17571 & 2270 & 599600 & 20240 \\
        +Gazz& 13269& 22211	& 2815 & 599600	& 20240 \\
        Gold (Test) & 569 & 510 & 353 & 14427 & 737 \\
        Clean 1.2K & 1068 & 1773 & 720 & 38423 & 1220 \\
        \bottomrule
    \end{tabular}
    \label{tab:dataset-stats}
}
\hfill
\parbox{.45\linewidth}{
    \centering
    \footnotesize
    \caption{1.2K instances of silver annotation performance with respect to the Clean 1.2k annotation. Clean 1.2k annotation is \textbf{subset} of DEE, MDEE and +Gazz}
    \begin{tabular*}{0.45\textwidth}{l @{\extracolsep{\fill}} |c|c|c@{}}
        \toprule
        Annotation & Prec & Recall & F1 \\
        \midrule
        DEE (1.2K) & 60.85 & 33.08 & 42.86 \\
        MDEE (1.2K) & 61.77 & 35.07 & 44.74 \\
        +Gazz (1.2K) & 63.83 & 40.44 & 49.51 \\
        \bottomrule
    \end{tabular*}
    \label{tab:silver-data}
}
\end{table}

\subsubsection{DBPedia Entity Expansion}

Our research used publicly available DBPedia Entity Expansion (DEE, Gold) \cite{Alfina2016DBpediaEE} and Modified Rule (MDEE, +Gazetteers) \cite{Alfina2017ModifiedDE} dataset for Indonesian. Interested readers should check the original references for further details. The dataset label statistics can be found in Table \ref{tab:dataset-stats}. We used the same test (Gold) in silver annotation Indonesian NER dataset. However, due to entity expansion technique, previous works \cite{Alfina2016DBpediaEE, Alfina2017ModifiedDE} only considers Entity without their span (BIO) labels. In order to alleviate this difference, we transform the contiguous Entity with same label into BIO span. This rule based conversion does not seem affecting exact match span-based F1-metrics in distantly supervised scenarios when we reproduce the model in the same configuration.


\subsection{ID-POS Corpus} The ID-POS corpus \cite{Rashel2014BuildingAI} contains 10K sentences of 250K tokens from news domain. There are 23 labels in the dataset. For POS tagging model, we train 5 model of 5-fold cross-validation following split dataset by \cite{kurniawan2018standardized}. For each fold of the models, we transfer the pre-trained weights into all NER train dataset in both large distantly supervised and low-resources gold NER scenarios.

\subsection{Unlabeled Corpus for Language Model} Total number of vocabulary in Wikipedia Indonesia are 100k unique tokens from 2 millions total sentences with 82 millions total tokens. While total number of vocabulary in Kompas \& Tempo dataset \cite{tala2003study} are 130k tokens from 85k total sentences with 11 millions total tokens.

\section{Experiments}
Our main experiment for cross-lingual settings is Austronesian language, Indonesian. We choose Indonesian due to its language characteristics such as morphological distance from Indo-European family but same Latin alphabet orthography to English. It contains many loanwords for verb and named entity words from several languages. Most of the named entity are kept in the same form as the original language lexicon. It also categorized as low-resources as there is no large scale standardized and publicly available gold annotated dataset for NER task.

We use AllenNLP \cite{Gardner2017AllenNLP} implementation for Baseline BiLSTM-CRF and extend our own implementation based on Supervised Cross-lingual Transfer, Cross-lingual using ELMo from EN, Monolingual ELMo and Unsupervised-Supervised Cross-lingual Transfer. We make our extension and pre-trained bi-LM of mono-lingual and cross-lingual available on \href{}{Github Links (Anonymous)}. We do not tune the model hyper-parameter such as dropout or learning rate, as there is no gold validation on comparable scenario with \cite{Alfina2017ModifiedDE}. In addition, we found that tuning hyper-parameter to noisy validation do not improve and can even lead to worse result such as over-fitting to false negative.

\subsubsection{General Model Configuration}

 We initialize all NER neural models on both monolingual and cross-lingual of Indonesian as target by using pre-trained word embedding with Glove \cite{D14-1162} on our Wikipedia dumps. The Glove-ID vectors are freeze during training on DEE, MDEE and +Gazz data. All the Indonesian NER models on distantly supervised data are trained for 10 epochs using Adam \cite{Kingma2014AdamAM} with learning rate $0.001$ for Optimization of batch size 32. For model using ELMo module, we use dropout rate 0.5 after the last layer output and before concatenation with word embedding and l2 regularization \cite{Krogh:1991:SWD:2986916.2987033} on ELMo weights to prevent model over-fitting and retain pre-trained knowledge. We use 2 layer Bi-LSTM-CRF layer with hidden size 200 and the word embedding dimension 50.


\subsubsection{Unsupervised Cross-lingual NER Transfer via ELMo}
In cross-lingual bi-directional LM using \texttt{CL via ELMo EN scenario}, we use pre-trained weights from English 1B tokens\footnote{\href{https://s3-us-west-2.amazonaws.com/allennlp/models/elmo/2x4096_512_2048cnn_2xhighway_tf_checkpoint/checkpoint}{model-checkpoint}} to Indonesian News dataset (IDNews) \cite{tala2003study}. We use implementation of bidirectional language model by Peters et al., (2018) \cite{P17-1161, Peters:2018} \footnote{https://github.com/allenai/bilm-tf} and modified it for cross-lingual transfer scenario. We fine-tune the model for 3 epochs by replacing the Softmax vocab layer with randomly initialized weight. We only fine-tune language model in cross-lingual scenarios on 3 epochs instead of 10 is to prevent catastrophic forgetting \cite{Robins1995CatastrophicFR}, \cite{P18-1031}.We called this model \texttt{ELMo EN-ID Transfer}. As a baseline, we use \texttt{ELMo EN-1B Tokens model} directly in the \texttt{CL via ELMo EN} scenario.


\begin{figure*}[htbp]
  \centering
  \includegraphics[width=1.02\textwidth]{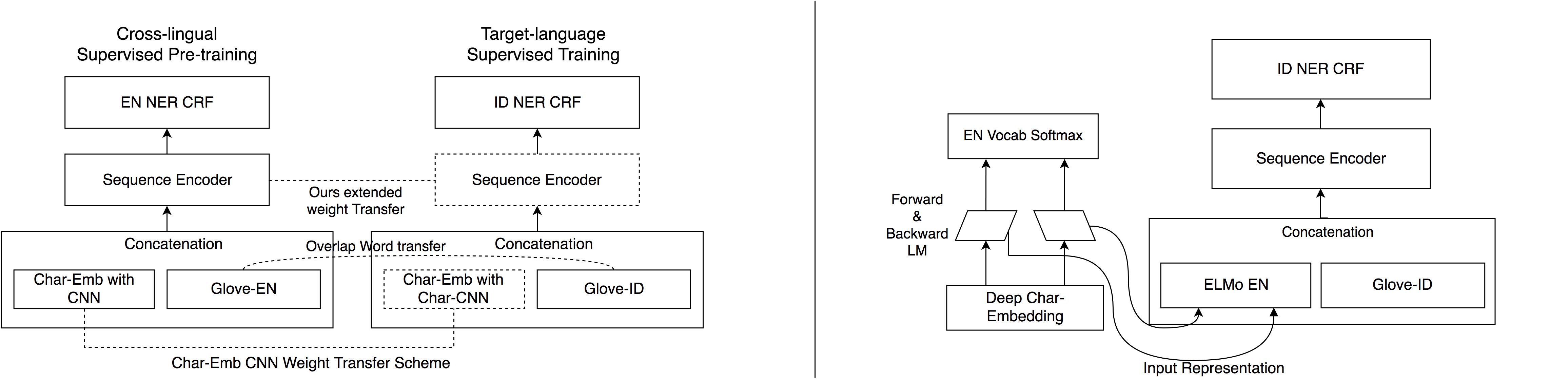}
  \caption{Left image, Baseline scenario for supervised cross-lingual transfer learning. Right image, Baseline scenario for directly using ELMo 1B Tokens EN initializer}
  \label{fig:base-transfer}
\end{figure*}


\subsubsection{Supervised Cross-lingual NER Transfer}\label{subsub:su-cl}

For the cross-lingual transfer learning baseline scenario, we use WP2, WP3 \cite{U08-1016} and CoNLL 2003 dataset \cite{TjongKimSang} of English language to train standard BiLSTM-CRF without ELMo initializer on 1B Language Model benchmarks. The models are trained on English languages and then the pre-trained weights are used as initalizer for both supervised and unsupervised transfer learning on DEE, MDEE, and +Gazz dataset. For the pre-trained English model, we report our reproduced baseline, recent state-of-the-arts NER and ELMo LSTM-CRF on WikiNER dataset \cite{U08-1016} to show the improvement on noisy mono-lingual data and use as pre-trained model. We train the English NER models for 75 epochs with patience 25 epochs for early stopping during training. In the experiment result Table \ref{tab:exp-result}, the model corresponds to [Sources] \texttt{BiLSTM-CRF} in "\texttt{Supervised CL NER Transfer}" scenarios.

\subsubsection{Mono-lingual ELMo}
In this scenarios, we use directly Pre-trained bi-LM on a mono-lingual corpus such as 1 billions word English \cite{chelba2013billion}, 82 millions Indonesian Wikipedia or 11 millions Indonesian News \cite{tala2003study} dataset which illustrated on Figure \ref{fig:base-transfer} on the right. In the experiment result Table \ref{tab:exp-result}, the model corresponds to \texttt{ELMo} ([Unlabeled corpus]) in "\texttt{Mono-lingual ELMo}"

\subsubsection{POS Tagging Transfer}
In this scenarios, we train a standard Bi-LSTM model using Softmax with Cross-entropy loss function to Indonesian POS tagging dataset. The transfer procedure almost the same as Supervised Cross-lingual NER Transfer as illustrated in Figure \ref{fig:base-transfer} on the right, while there are 2 differences i) the top-most layer is Linear with Softmax Activation instead of CRF, and ii) the sources task is POS tagging instead of English NER. We train 5 models based on 5-fold cross-validation split provided by Kurniawan et al., (2018) \cite{kurniawan2018standardized}, we report the averaged F1 of each k-th-fold model as pre-trained weights in both large silver and small clean annotation. In the experiment result Table \ref{tab:exp-result}, the model corresponds to \texttt{ID-POS BiLSTM-CRF} in "\texttt{POS Tagging Transfer}" scenario. 

This experiment scenario serve as comparison of transfer learning from different but related task in Yang et al., (2017) \cite{Yang2016TransferLF}. In addition, previous work by Blevins et al. (2018) \cite{P18-2003} show that LM contains syntactic information thus serve as comparison to pre-trained monolingual bidirectional LM.

\subsubsection{Multi-Task NER with BiLM} 
We also train and evaluate using recent state-of-the-arts model in Indonesian conversational dataset such as Multi-Task NER with BiLM  auxiliary task (BiLM-NER) \cite{DBLP:journals/corr/abs-1805-12291}. In the experiment Table \ref{tab:exp-result}, the model corresponds to \texttt{BiLM-NER} in "\texttt{Baseline}" scenarios.

\section{Results \& Analysis}
In this research, we reports our English dataset results which mainly used to show improvement of pre-trained BiLM and as source weights in transfer learning. We reports our main experiments in several version of large silver for model comparison and a small clean annotation in ablation scenarios. Finally, we analyzed our proposed method of supervised cross-lingual transfer with BiLM and Cross-lingual Transfer via Language Model.

\subsection{English Dataset Results} From Table \ref{tab:distant}, model trained using pre-trained ELMo and random Word Embedding initialization (WE+ELMo LSTM-CRF) are better with an average of 4.925 \% F1 score in four WikiNER scenarios compare to Word embedding initialized with Glove 6B words and character-CNN (WE+CharEmb) on CoNLL dataset. However, it is tie on WikiGold test where Glove+CharEmb without MISC labels perform are better than WE+ELMo, whereas the latter are better with MISC labels than the former. Overall, combining both Glove and ELMo yields best results except when using WP2 as training data when tested in CoNLL test.

\begin{table}
    \scriptsize
    \centering
    \caption{F1 score performance results on WikiGold and CoNLL test set. English NER model w/o (without) MISC and pre-trained weight Glove 6B \& ELMo 1B used as pre-train model for cross-lingual transfer scenarios}
    \label{tab:distant}
    \begin{tabular}{@{}l |c|c|c @{}}
        \toprule
         Train Data &  WikiGold & CoNLL & Pre-Init\\
         \midrule
          \multicolumn{4}{c}{\multirow{-1}{*}{Glove+CharEmb LSTM-CRF}} \\
          \midrule
        WP2 & 71.75 & 61.78 & Glove 6B\\
        WP3 & 71.40 & 62.51 & Glove 6B\\
        CoNLL & 58.00 & 90.47 & Glove 6B\\
        \midrule
        WP2-w/o MISC & 75.12 & 65.35 & Glove 6B\\
        WP3-w/o MISC & 75.02 & 63.69 & Glove 6B\\
        CoNLL-w/o MISC & 58.30 & 91.37 & Glove 6B\\
        \midrule
        \multicolumn{4}{c}{\multirow{-1}{*}{WE (Random Init) +ELMo LSTM-CRF}} \\
        \midrule
        WP2 & 76.96 & \textbf{71.48} & ELMo 1B\\
        WP3 & 74.95 & 68.54 & ELMo 1B\\
        CoNLL & 74.07 & 90.18 & ELMo 1B\\
        \midrule
        WP2-w/o MISC & 73.47 & 66.50 & ELMo 1B\\
        WP3-w/o MISC & 72.91 & 66.51 & ELMo 1B\\
        CoNLL-w/o MISC & 74.52 & 91.59 & ELMo 1B\\
        \midrule
        \multicolumn{4}{c}{\multirow{-1}{*}{Glove +ELMo LSTM-CRF}} \\
        \midrule
        WP2 & \textbf{77.14} & 69.91 & Glove 6B \& ELMo 1B\\
        WP3 & \textbf{76.92} & \textbf{70.31} & Glove 6B \& ELMo 1B\\
        CoNLL & \textbf{75.12} & \textbf{91.98} & Glove 6B \& ELMo 1B\\
        \midrule
        WP2-w/o MISC & \textbf{80.55} & \textbf{73.05} & Glove 6B \& ELMo 1B\\
        WP3-w/o MISC & \textbf{81.09} & \textbf{75.60} & Glove 6B \& ELMo 1B\\
        CoNLL-w/o MISC & \textbf{79.49} & \textbf{93.53} & Glove 6B \& ELMo 1B\\
        \bottomrule
    \end{tabular}
\end{table}

\subsection{Indonesian Dataset Results}

\begin{table}
\parbox{.5\linewidth}{
    \scriptsize
    \centering
    \caption{Experiment on silver standard annotation of Indonesian NER \textbf{evaluated on Gold test set} \cite{Alfina2016DBpediaEE} in \textbf{large distantly supervised NER scenario}. Bold F1 scores are best result per scenarios (Baseline, Supervised Cross-lingual Transfer, Cross-lingual using ELMo from EN, Mono-lingual ELMo and Unsupervised-Supervised Cross-lingual Transfer). * is the best model on a dataset (DEE, MDEE, or +Gazz) on all model scenarios}
    \label{tab:exp-result}
    \begin{tabular*}{0.5\textwidth}{l @{\extracolsep{\fill}}|c|c|c}
        \toprule
        Model & DEE & MDEE & +Gazz \\
         \midrule
         \multicolumn{4}{c}{Previous Works} \\
         \midrule
         Alfina et al., \cite{Alfina2017ModifiedDE} & 41.33 & 41.87 & 51.61 \\
         BiLM-NER & 40.36 & 41.03 & 51.77 \\
         \midrule
         \multicolumn{4}{c}{Baseline} \\
         \midrule
         Stanford-NER-BIO \cite{Alfina2017ModifiedDE} & 40.68 & 41.17 & 51.01 \\
         BiLSTM-CRF & \textbf{46.09} & \textbf{45.59} & \textbf{52.04} \\
        \midrule
         \multicolumn{4}{c}{POS Tagging Transfer} \\
        \midrule
         ID-POS BiLSTM-CRF & 52.58 & 51.07 & 60.57 \\
         \midrule
         \multicolumn{4}{c}{Supervised CL NER Transfer} \\
         \midrule
        WP2 BiLSTM-CRF & 49.88 & \textbf{52.35} & 62.57 \\
        WP3 BiLSTM-CRF & 51.21 & 50.95 & \textbf{62.90} \\
        CoNLL BiLSTM-CRF & \textbf{52.56} & 50.75 & 60.81 \\
         \midrule
         \multicolumn{4}{c}{CL via ELMo EN} \\
         \midrule
         ELMo EN-1B Tokens & 51.08 & 53.19 & 60.66 \\
         ELMo EN-ID Transfer & \textbf{52.63} & \textbf{54.74} & \textbf{63.02} \\
         \midrule
         \multicolumn{4}{c}{Mono-lingual ELMo} \\
         \midrule
        ELMo (ID-Wiki) & \textbf{50.68} & \textbf{52.38} & 60.51 \\
         ELMo (ID-News) & 49.49 & 51.91 & \textbf{60.73} \\
         \midrule
         \multicolumn{4}{c}{Supervised CL Transfer with ELMo} \\
         \midrule
        WP2 ELMo (EN) & 52.99 & \textbf{55.39}* & 63.99 \\
        WP3 ELMo (EN) & \textbf{54.15}* & 55.28 & 63.84 \\
        CoNLL ELMo (EN) & 53.52 & 53.48 & \textbf{64.35}* \\
        \bottomrule
    \end{tabular*}
}
\hfill
\parbox{.45\linewidth}{
    \scriptsize
    \centering
    \caption{Ablation experiment results using Clean 1.2K as training data in \textbf{small clean (human annotated) scenario} also \textbf{evaluated on Gold test set}. W: Word embedding (Random Init), C: Char-CNN (+EN if INIT from CoNLL 2003) embedding, E: ELMo (EN), G: Glove-ID(+EN if in cross-lingual transfer from English) \cite{D14-1162}, I: ELMo (ID-Wiki),  J: ELMo (EN-ID-News) Transfer}
    \label{tab:ablation-result}
    \begin{tabular*}{0.45\textwidth}{l @{\extracolsep{\fill}} | c| c| c@{}}
        \toprule
         Model & Prec & Rec & F1\\
         \midrule
         Stanford-NER & 71.42 & 53.84 & 61.39\\
         \midrule
         BiLM-NER & 63.65 & 63.29 & 63.47\\
         \midrule
         \multicolumn{4}{c}{BiLSTM-CRF}\\
         \midrule
        W+C+E & 76.42 & 56.32 & 64.85\\
         W+C & 56.23 & 56.39 & 56.31\\
         W+E & 73.53 & 53.32 & 61.81\\
         C+E & 69.13 & 68.60 & 68.86\\
         G & 63.65 & 48.50 & 55.05\\
         G+C & 69.17 & 62.31 & 65.56\\
         G+E & 75.30 & 65.32 & 69.96\\
         G+C+E & 72.05 & 68.73 & 70.35\\
         E & 76.27 & 55.41 & 64.19\\
         G+C+I & 74.53 & 78.43 & 76.43\\
         G+I & 75.57 & 77.94 & 76.74\\
         I & 78.55 & 73.62 & 76.00\\
         G+C+J & 83.26 & 82.62 & 82.94\\
         G+J & 83.77 & 83.60 & \textbf{83.68}\\
         J & 82.36 & 83.74 & 83.04\\
         \midrule
         \multicolumn{4}{c}{ INIT from ID-POS }\\
         \midrule
         W+C & 72.97 &  78.97 & 75.68\\
         \midrule
         \multicolumn{4}{c}{ INIT from CoNLL 2003 }\\
         \midrule
         W+C & 66.23 & 56.25 & 60.83\\
         G+C & 70.18 & 65.87 & 67.96\\
         C+E & 71.84 & 64.27 & 67.85\\
         W+C+E & 73.63 & 65.46 & 69.30\\
         G+E & 73.38 & 69.08 & 71.17\\
         G+C+E & 72.63 & 72.99 & 72.85\\
        \bottomrule
    \end{tabular*}
}
\end{table}

We reproduce around the same results of \cite{Alfina2017ModifiedDE} using Stanford NER. Our experiment using a recent state-of-the-arts model in Indonesian conversational dataset namely Multi-Task NER with BiLM auxiliary task (BiLM-NER) \cite{DBLP:journals/corr/abs-1805-12291} (\texttt{BiLM-NER}) obtain comparable performance with log-linear model but lower than BiLSTM-CRF \cite{P16-1101}.

The mono-lingual pre-trained BiLM on 1B English words (ELMO EN-1B Tokens) performs comparable with pre-trained BiLM on 82 millions tokens in (\texttt{ELMo} (ID-Wiki)) and 11 millions news tokens (\texttt{ELMo} (ID-News)). All of the mono-lingual Embedding from Pre-trained BiLM on silver standard annotation perform worse than baseline supervised cross-lingual with \& without BiLM scenarios.

\subsection{Cross-lingual Transfer Analysis}

We hypotheses that the performance of using ELMo on cross-lingual settings despite a little counter-intuitive are not entirely surprising can be addressed to i) Most named entities which available on multi-lingual documents are orthographically similar. For instance "America" is "\emph{Amerika}" in Indonesian, while "Obama" is still "\emph{Obama}", "President Barack Obama" is still "\emph{Presiden} Barack Obama"; ii) Due to the orthographic similarities of many entity names, the fact that English and Indonesian languages are typologically different (e.g. in terms of S-V-O word order and Determiner-Noun word order) is not relevant on noisy data, as long as the character sequences of named entities are similar in both languages \cite{I17-2016, D18-1034}.

We confirm our first hypothesis by looking up the percentage of unique word (vocabulary) overlap rate between the Gold ID-NER \cite{Alfina2016DBpediaEE} and three English dataset, namely WP2, WP3 \cite{U08-1016} and CoNLL training \cite{TjongKimSang}. The overall vocabulary overlap rate between Gold ID-NER and the three dataset are $26.77\%$, $25.70\%$, $15.24\%$ respectively. Furthermore, we checked WP2 per word-tag join overlap rate are \texttt{PER} $51.09\%$, \texttt{LOC} $60.9\%$, \texttt{ORG} $60.54\%$, and \texttt{O} $16.56\%$ percentage. While CoNLL word-tag joins overlap rate are \texttt{PER} $37.53\%$, \texttt{LOC} $27.54\%$, \texttt{ORG} $39.46\%$, and \texttt{O} $9.23\%$. More details of unique word overlap rate between Indonesian DBPedia Entity, WP2, WP3 and CoNLL can be seen on Table \ref{tab:exp-result}. in Supervised Cross-lingual Transfer which only utilized character-embedding and pre-trained monolingual word-embedding trained from CoNLL dataset perform worse on both MDEE and +Gazz dataset than trained on WP2 and WP3 dataset. 

We support our second hypothesis by doing ablation on clean annotation (Table \ref{tab:ablation-result}). Our clean annotation show that, ELMo (ID-Wiki) outperformed ELMo (EN-1B Tokens) on small clean annotation data, but ELMo EN nonetheless still outperformed BiLSTM-CRF especially when combined with Supervised pre-training on CoNLL 2003 English NER \cite{P16-1101}.

\begin{figure}[htbp]
  \centering
  \includegraphics[width=0.75\textwidth]{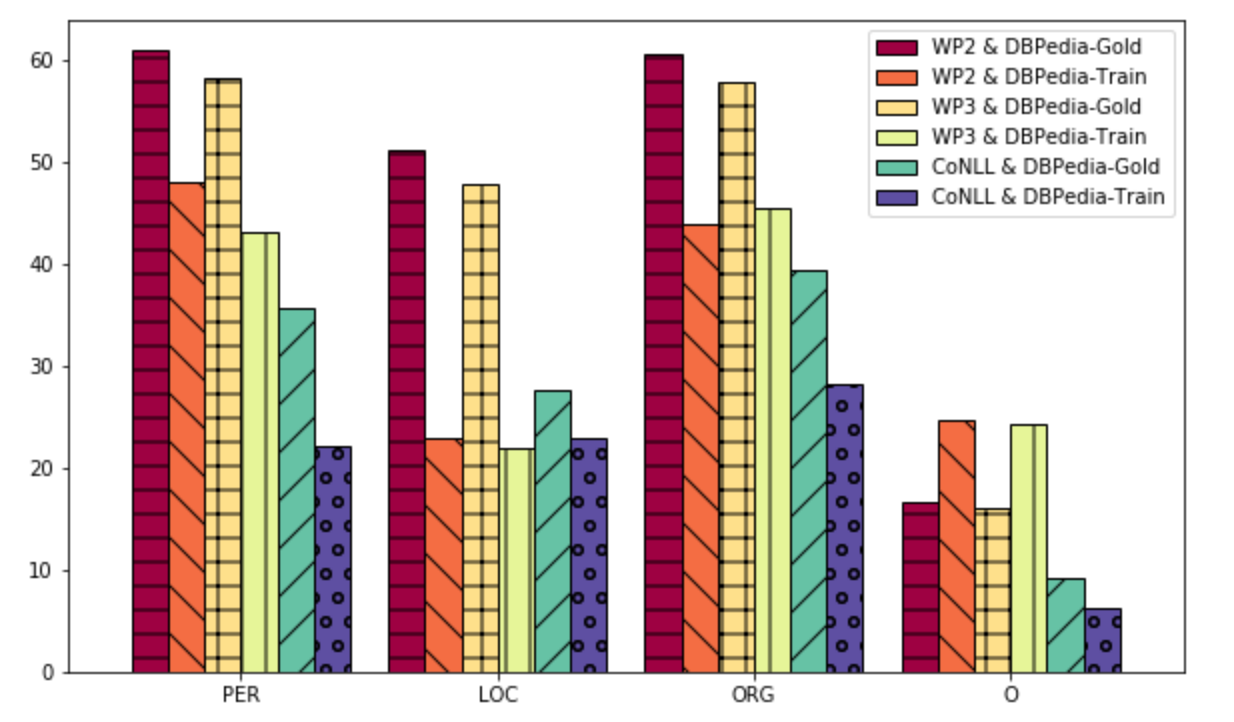}
  \caption{Word-tag overlap rate breakdown between mono-lingual and cross-lingual corpora. (-) horizontal line: WP2 \& DBPedia Gold, right slope: WP2 \& DBPedia Train, (+) cross: is overlap between WP3 \& DBPedia Gold, (|) vertical: overlap between WP3 \& DBPedia Train, (/) left slope: CoNLL Train and DBPedia Gold, (o) dot: CoNLL Train and DBPeida Train}
  \label{fig:cl-word-overlap}
\end{figure}

\section{Conclusion}

In this research, we extend the idea of character-level embedding pre-trained on language model to cross-lingual scenarios for distantly supervised and low-resources scenarios. We observed that training character-level embedding of language model requires enormous size of corpora \cite{Peters:2018}. Addressing this problem, we demonstrate that as long as orthographic constraint and some lexical words in target language such as loanwords to act as pivot are shared, we can utilize the high-resource languages model.

\section*{Acknowledgments}

    We  also  would  like  to  thank Samuel Louvan, Kemal Kurniawan, Adhiguna Kuncoro, and Rezka Aufar L. for reviewing the early version of this work. We are also grateful to Suci Brooks and 
    Pria  Purnama for their relentless support. \\

\bibliographystyle{splncs04}
\bibliography{refs}





\end{document}